\documentclass[pdflatex,sn-mathphys-num]{sn-jnl}

\usepackage{graphicx}
\usepackage{multirow}
\usepackage{amsmath,amssymb,amsfonts}
\usepackage{amsthm}
\usepackage{mathrsfs}
\usepackage[title]{appendix}
\usepackage{xcolor}
\usepackage{textcomp}
\usepackage{manyfoot}
\usepackage{booktabs}
\usepackage{algorithm}
\usepackage{algorithmicx}
\usepackage{algpseudocode}
\usepackage{listings}
\usepackage{float}
\usepackage{array}
\usepackage{adjustbox}
\usepackage{dsfont}
\usepackage{subcaption}
\usepackage{multibib}
\usepackage{tabularx}
\usepackage{paracol}
\usepackage{stfloats}
\usepackage{placeins}
\usepackage{mathtools}
\usepackage{physics}
\usepackage{nicefrac}
\usepackage{microtype}
\usepackage{enumitem}
\usepackage{thmtools,thm-restate}
\usepackage{bm}
\usepackage{comment}
\usepackage{ragged2e}
\usepackage{hyperref}
\usepackage{tabularx}
\usepackage{rotating}
\usepackage{makecell} 
\usepackage{setspace}

\theoremstyle{thmstyleone}%
\newtheorem{theorem}{Theorem}
\theoremstyle{thmstyletwo}%

\theoremstyle{thmstylethree}%

\newtheorem{lemma}[theorem]{Lemma}
\newtheorem{prop}[theorem]{Proposition}

\DeclareMathOperator{\cA}{\mathcal{A}}
\DeclareMathOperator{\cS}{\mathcal{S}}
\DeclareMathOperator{\cD}{\mathcal{D}}
\DeclareMathOperator{\cM}{\mathcal{M}}
\DeclareMathOperator{\cR}{\mathcal{R}}
\DeclareMathOperator{\cX}{\mathcal{X}}
\DeclareMathOperator{\cF}{\mathcal{F}}
\DeclareMathOperator{\cH}{\mathcal{H}}

\usepackage{amsmath,amsfonts,bm}

\def\eqref#1{equation~\ref{#1}}

\def\1{\bm{1}}

\DeclareMathAlphabet{\mathsfit}{\encodingdefault}{\sfdefault}{m}{sl}
\SetMathAlphabet{\mathsfit}{bold}{\encodingdefault}{\sfdefault}{bx}{n}

\newcommand{\pch}[1]{\textcolor{black}{#1}}
\newcommand{\revECML}[1]{\textcolor{black}{#1}}

\newcommand{\todo}[1]{\textcolor{black}{#1}}
\newcommand{\ie}{\textit{i}.\textit{e}.,\ }
\newcommand{\eg}{\textit{e}.\textit{g}.,\ }
\makeatletter
\newcommand{\multiline}[1]{%
  \begin{tabularx}{\dimexpr\linewidth-\ALG@thistlm}[t]{@{}X@{}}
    #1
  \end{tabularx}
}

\begin{document}

\title[Enhancing Offline Model-Based RL via Active Model Selection: A Bayesian Optimization Perspective]{Enhancing Offline Model-Based RL via Active Model Selection: A Bayesian Optimization Perspective}

\author[1]{Yu-Wei Yang}\email{ray17606@gmail.com}
\author[1]{Wei Hung}\email{hwei1048576.cs08@nycu.edu.tw}
\author[1]{Yun-Ming Chan}\email{yunming.cs11@nycu.edu.tw}

\author[2]{Xi Liu}\email{xiliu.tamu@gmail.com}
\author*[1]{Ping-Chun Hsieh}\email{pinghsieh@nycu.edu.tw}

\affil[1]{Department of Computer Science, National Yang Ming Chiao Tung University, Hsinchu, Taiwan}
\affil[2]{Applied Machine Learning, Meta AI, Menlo Park, CA, USA}


\abstract{Offline model-based reinforcement learning (MBRL) serves as a competitive framework that can learn well-performing policies solely from pre-collected data with the help of learned dynamics models. To fully unleash the power of offline MBRL, model selection plays a pivotal role in determining the dynamics model utilized for downstream policy learning. However, offline MBRL conventionally relies on validation or off-policy evaluation, which are rather inaccurate due to the inherent distribution shift in offline RL. To tackle this, we propose BOMS, an active model selection framework that enhances model selection in offline MBRL with only a small online interaction budget, through the lens of Bayesian optimization (BO). Specifically, we recast model selection as BO and enable probabilistic inference in BOMS by proposing a novel model-induced kernel, which is theoretically grounded and computationally efficient. Through extensive experiments, we show that BOMS improves over the baseline methods with a small amount of online interaction comparable to only $1\%$-$2.5\%$ of offline training data on various RL tasks.}


\maketitle



\section{Introduction}
\label{sec:intro}
In conventional online reinforcement learning (RL), agents learn through direct online interactions, posing challenges in real-world scenarios where online data collection is either costly (\eg robotics \citep{kalashnikov2018scalable, ibarz2021train} and recommender systems \citep{afsar2022reinforcement, zhao2021dear} under the risk of damaging machines or losing customers) or even dangerous (\eg healthcare \citep{yu2021reinforcement, yom2017encouraging}). \textit{Offline RL} addresses this challenge by learning from pre-collected datasets, showcasing its potential in applications where online interaction is infeasible or unsafe.
To enable offline RL, existing methods mostly focus on addressing the fundamental \textit{distribution shift} issue and avoid visiting the out-of-distribution state-action pairs by encoding conservatism into the policy \citep{levine2020offline,kumar2020conservative}.

As one major category of offline RL, offline \textit{model-based} RL (MBRL) implements such conservatism through one or an ensemble of dynamics models, which is learned from the offline dataset and later modified towards conservatism, either by uncertainty estimation \citep{yu2020mopo, kidambi2020morel,yu2021combo,rashidinejad2021bridging,sun2023model} or adversarial training \citep{rigter2022rambo,bhardwaj2023adversarial}. 
The modified dynamics models facilitate the policy learning by offering additional imaginary rollouts, which could achieve better data efficiency and potentially better generalization to those states not present in the offline dataset. Moreover, several offline MBRL approaches \citep{rigter2022rambo,sun2023model,bhardwaj2023adversarial} have achieved competitive results on the D4RL benchmark tasks \citep{fu2020d4rl}.

Despite the progress, one fundamental and yet underexplored issue in offline MBRL lies in the \textit{selection of the learned dynamics models}. Specifically, in the phase of learning the transition dynamics, one needs to determine either a stopping criterion or which model(s) in the candidate model set to choose and introduce conservatism to for the downstream policy learning. Notably, the model selection problem is challenging in offline RL for two reasons: (i) \textit{Distribution shift}: Due to the inherent distribution shift, the validation of either the dynamics models or the resulting policies could be quite inaccurate; (ii) \textit{Sensitivity of neural function approximators}: To capture the complex transition dynamics, offline MBRL typically uses neural networks as parameterized function approximators, but the neural networks are widely known to be highly sensitive to the small variations in the weight parameters \citep{jacot2018neural,tsai2021formalizing, liu2017fault}. 
As a result, it has been widely observed that the learning curves of offline MBRL can be fairly unstable throughout training, as pointed out by \citep{lu2021revisiting}.

\vspace{2mm}
\noindent\textbf{Limitations of the existing model selection schemes:} The existing offline MBRL methods typically rely on one of the following model selection schemes: 

\begin{itemize}[leftmargin=*]
    \item \textit{Validation as in supervised learning}: As the learning of dynamics models could be viewed as an instance of supervised learning, model selection could be done based on a validation loss, \eg mean squared error with respect to a validation dataset \citep{janner2019trust,yu2020mopo}. However, as the validation dataset only covers a small portion of the state and action spaces, the distribution shift issue still exists and thereby leads to poor generalization.
    \item \textit{Off-policy evaluation}: Another commonly-used approach is off-policy evaluation (OPE) \citep{paine2020hyperparameter,voloshin2019empirical}, where given multiple candidate dynamics models, one would first learn the corresponding policy based on each individual model, and then proceed to choose the policy with the highest OPE score. However, it is known that the accuracy of OPE is limited by the coverage of the offline dataset and therefore could still suffer from the distribution shift \citep{tang2021model}. 
\end{itemize}
To further illustrate the limitations of the validation-based and OPE-based model selection schemes, we take the model selection problem of MOPO \citep{yu2020mopo}, a popular offline model-based RL algorithm, as a motivating example and empirically evaluate these two model selection schemes on multiple MuJoCo locomotion tasks of the D4RL benchmark datasets \citep{fu2020d4rl}. Please refer to Appendix \ref{app:config} for the experimental configuration. 
Table~\ref{table:motivating} shows that the total rewards achieved by the two model selection schemes can be much worse than that of the best choice (among 150 candidate models obtained via MOPO \citep{yu2020mopo}) and thereby demonstrates the significant sub-optimality of the existing model selection schemes.

\noindent\textbf{Active model selection:} \todo{Despite the difficulties mentioned above, in various practical applications, a small amount of online data collection is usually allowed to ensure reliable deployment \citep{lee2022offline, zhao2022adaptive, konyushova2021active}, such as robotics (testing of control software on real hardware), recommender systems (the standard A/B testing), and transportation networks (testing of the traffic signal control policies on demonstration sites). In this paper, we propose a new problem setting -- \textit{Active Model Selection}, where the goal is to reliably select models with the help of a small budget of online interactions (\eg $5\%$ of training data). Specifically, we answer the following important and practical research question:}

\textit{How to achieve effective model selection with only a small budget of online interaction for offline MBRL?}

\begin{table}[!t]
\centering
\caption{Total rewards achieved by MOPO~\citep{yu2020mopo} under validation-based and OPE-based model selection on D4RL. The performance of the best model is also reported to highlight the sub-optimality of the existing schemes.}
\begin{tabularx}{\textwidth}{lccc}
\toprule
           & \textbf{hopper (medium)} & \textbf{walker2d (medium)} & \textbf{halfcheetah (medium)} \\ \midrule
Best model & 2073 {\small$\pm 822$} & 4002 {\small$\pm 238$} & 6607 {\small$\pm 295$} \\
Validation & 821 {\small$\pm 440$} & 39 {\small$\pm 61$} & 5978 {\small$\pm 140$} \\
OPE        & 495 {\small$\pm 265$} & 3618 {\small$\pm 241$} & 4355 {\small$\pm 581$} \\ \bottomrule
\end{tabularx}
\label{table:motivating}
\end{table}

\begin{figure}[!ht]
    \centering
    \includegraphics[width=0.45\textwidth]{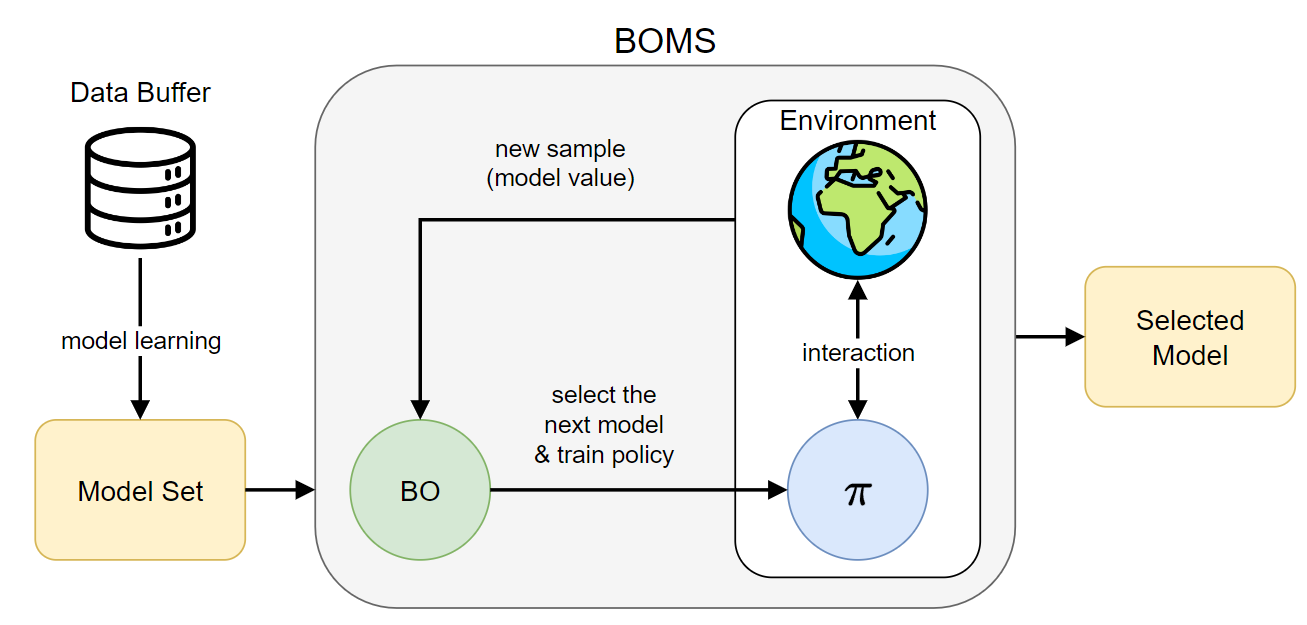}
    \caption{An illustration of the BOMS framework.}
    \label{fig:BOMS}
\end{figure}

\vspace{2mm}
\noindent\textbf{Bayesian optimization for model selection:} In this paper, we propose to address \textit{Active Model Selection} for offline MBRL from a Bayesian optimization (BO) perspective. Our method aims to identify the best among the learned dynamics models with only a small budget of online interaction for downstream policy learning. In BO, the goal is to optimize an expensive-to-evaluate black-box objective function by iteratively sampling a small number of candidate points from the candidate input domain. Inspired by \citep{konyushova2021active}, we propose to recast model selection in offline MBRL as a BO problem (termed BOMS throughout this paper), where the candidates are the dynamics models learned during training and the objective function is the total expected return of the policy that corresponds to each dynamics model. BOMS proceeds iteratively via the following two steps: In each iteration $t$, (i) we first learn a surrogate function with the help of a Gaussian process (GP) and determine the next model to evaluate (denoted by $M_t$) based on the posterior under GP. (ii) We then evaluate the policy obtained under $M_t$ with a small number of online trajectories. 
Figure \ref{fig:BOMS} provides a high-level illustration of the BOMS framework. \revECML{Notably, this setting is distinct from strictly offline RL: BOMS is applicable when a small, controlled online-evaluation budget is available.}

\vspace{2mm}
\noindent\textbf{Model-induced kernels for BOMS:} In BOMS, one fundamental challenge is to determine the similarity of dynamics models (analogous to the Euclidean distances between domain points in standard BO). To enable BOMS, we propose a novel 
\textit{model-induced kernel function} that captures the similarity between dynamics models, and the proposed kernel is both computationally efficient and theoretically-grounded by the derived sub-optimality lower bound. 
Somewhat surprisingly, through extensive experiments, we show that significant improvement can be achieved through active model selection of BOMS with only a relatively small number of online interactions (\eg around 20 to 50 trajectories for various locomotion and robot manipulation tasks).

\section{Preliminaries}
\label{sec:prelim}
\textbf{Markov decision processes}. We consider a Markov decision processes (MDP) specified by a tuple $M:=(\cS, \cA, r, P, \omega, \gamma)$, where $\cS$ and $\cA$ denote the state space and action space, $r: \cS\times \cA \rightarrow [-r_{\max}, r_{\max}]$ is the reward function, $\omega$ is the initial state distribution, $P(s'\rvert s,a)$ denotes the transition kernel, and $\gamma\in [0,1)$ is the discount factor. Throughout this paper, we use $M^*$ to denote the MDP with the true transition dynamics and true reward function of the environment. We use $\pi$ to denote a stationary policy, which specifies the action distribution at each state. Let $\Pi$ denote the set of all stationary policies. 
For any $\pi \in \Pi$, the discounted state-action visitation distribution $\rho^\pi: \mathcal{S} \times \mathcal{A} \rightarrow \mathbb{R}$ is defined as that for each $(s,a)$,
\begin{equation}
   \rho^\pi(s,a):= (1-\gamma) \pi(a\rvert s) \sum_{t=0}^{\infty} \gamma^t P(s_t=s\rvert \pi).  
\end{equation}
Define the value function of a policy $\pi$ under $M^*$ as
\begin{equation}
V^\pi_{M^*}(s):= \displaystyle\mathop{\mathbb{E}}_{\substack{a_t\sim\pi(\cdot\rvert s_t)\\ s_{t+1}\sim P(\cdot\rvert s_t,a_t)}}\left[\sum_{t=0}^{\infty} \gamma^t r(s_t,a_t) \Big\rvert s_0=s \right].
\end{equation}
The performance of a policy $\pi$ is evaluated by its total expected discounted return, which is defined as
\begin{equation}
J^\pi_{M^*}(\omega):= \displaystyle\mathop{\mathbb{E}}_{s\sim\omega}\left[V^\pi(s)_M^*\right].
\end{equation}
The goal of RL is to learn an optimal policy $\pi^*$ that maximizes the total expected discounted return, \ie
\begin{equation}
  \pi^* \in \arg\max_{\pi\in \Pi} J^\pi_{M^*}(\omega).
\end{equation}

\noindent\textbf{Offline model-based RL.} In {offline MBRL} methods, the learner is required to learn a well-performing policy based solely on a fixed dataset denoted by $\cD_{\text{off}}:=\{(s_i,a_i,r_i,s_i')\}_{i=1}^{N}$, which is collected from the environment modeled by $M^*$ under some \textit{unknown} behavior policy $\pi_\beta$, without any online interaction.
For ease of notation, we also let $\hat{M}:=(\cS, \cA, \hat{r}, \hat{P}, \omega, \gamma)$ denote the MDP induced by the estimated dynamics model $\hat{P}$ and the estimated reward function $\hat{r}$.

For didactic purposes, we use the popular MOPO algorithm \citep{yu2020mopo} as an example to describe the common procedure of offline model-based methods. MOPO consists of three main subroutines as follows.

\begin{itemize}[leftmargin=*]
    \item \textit{Learn the dynamics models (along with some model selection scheme)}: MOPO learns the estimated $\hat{P}$ and $\hat{r}$ via supervised learning (\eg minimizing squared error) on the offline dataset $\cD_{\text{off}}$. Regarding model selection, MOPO uses a validation-based method that induces a stopping rule based on the validation loss.
    \item \textit{Construct an uncertainty-penalized MDP}: MOPO then construct a penalized MDP $\widetilde{M}:=(\cS, \cA, \widetilde{r}, \hat{P}, \omega, \gamma)$, where $\widetilde{r}$ is the penalized reward function defined as $\widetilde{r}(s,a):=\hat{r}(s,a)-\lambda u(s,a)$ with $u(s,a)$ being an uncertainty measure and $\lambda$ being the penalty weight. 
    \item \textit{Run an RL algorithm on uncertainty-penalized MDPs}: MOPO then learns a policy by running an off-the-shelf RL algorithm (\eg Soft Actor Critic \citep{haarnoja2018soft}) on the penalized MDP.
\end{itemize}
\noindent\textbf{Bayesian optimization.} 
As a popular framework of black-box optimization, BO finds a maximum point ${x^*}$ of an unknown and expensive-to-evaluate objective function $f(\cdot): \mathbb{R}^d \rightarrow \mathbb{R}$ through iterative sampling by incorporating two critical components:
\begin{itemize}[leftmargin=*]
\item \textit{Gaussian process (GP)}: To facilitate the inference of unknown black-box functions, BO leverages a GP function prior that characterizes the correlation among the function values and thereby implicitly captures the underlying smoothness of the functions. Specifically, a GP prior can be fully characterized by its mean function $m(\cdot): \mathcal{X}\rightarrow \mathbb{R}$ and the kernel function $k(\cdot,\cdot): \mathcal{X}\times \mathcal{X} \rightarrow \mathbb{R}$ that determines the covariance between the function values at any pair of points in the domain $\mathcal{X}$ \citep{schulz2018tutorial}. As a result, in each iteration $t$, given the first $t$ samples $\mathcal{H}_t:=\{(x_i,y_i)\}_{i=1}^{t}$, the posterior distribution of each $x$ in the domain remains a Gaussian distribution denoted as $\mathcal{N}(\mu_t(x),\sigma^2_t(x))$, where $\mu_t(x):=\mathbb{E}[f(x)\rvert \mathcal{H}_t]$ and $\sigma_t(x):=\sqrt{\mathbb{V}[f(x)\rvert \mathcal{H}_t]}$ could be derived exactly.
Notably, this is one major benefit of using a GP prior in black-box optimization.
\item \textit{Acquisition functions}: In general, the sampling process of BO could be viewed as solving a sequential decision making problem. However, applying any standard dynamic-programming-like method to determine the sampling could be impractical due to the curse of dimensionality \citep{wang2016bayesian}. To enable efficient sampling, BO typically employs an acquisition function (AF) denoted by $\Psi(x; \cF_t)$, which provides a ranking of the candidate points $x\in \cX$ in a myopic manner based on the posterior means and variances given the history $\cF_t$. As a result, an AF can induce a tractable index-type sampling policy. Existing AFs have been developed from various perspectives, such as the improvement-based methods (\eg Expected Improvement \citep{jones1998efficient}), optimism in the face of uncertainty (\eg, GP-UCB \citep{srinivas2009gaussian}), information-theoretic approaches (\eg MES \citep{wang2017max}), and the learning-based methods (\eg MetaBO \citep{volpp2019meta} and FSAF \citep{hsieh2021reinforced}).
\end{itemize}

\section{Methodology}
\label{sec:boms}

\subsection{Recasting Model Selection as BO}
\label{sec:boms:framework}
We start by connecting the active model selection problem to BO. Specifically, below we interpret each component of model selection in the language of BO.
\begin{itemize}[leftmargin=*]
    \item \textit{Input domain $\cX$}: Recall that offline model-based RL first uses the offline dataset $\cD_{\text{off}}$ to generate a collection of ${N}$ dynamics models denoted by $\cM$ ${= \{M^{(1)}, M^{(2)}, ..., M^{(N)}\}}$, which serve as the candidate set for model selection. We view this candidate set $\cM$ as the input domain in BO, \ie $\cM\equiv \cX$.
    \item \textit{Black-box objective function}: The goal of model selection is to choose a dynamics model from $\cM$ such that the policy learned downstream enjoys high expected return. Let $\pi^{(i)}$ be the policy learned under $M^{(i)}\in \cM$, and let $J_{M^*}^{\pi^{(i)}}$ be the true total expected return achieved by $\pi^{(i)}$. Accordingly, model selection can be viewed as black-box maximization with an unknown objective function $f \equiv J^{\pi}_{M^*}$ defined on $\cM$.
    \item \textit{Sampling procedure}: In the context of model selection, each sample corresponds to evaluating the policy learned from a model $M'\in \cM$. We let $M_t$ and $J^{\pi_t}_{M^*}$ denote the model selected and the corresponding policy performance at the $t$-th iteration, respectively. Let $\cH_t:=\{(M_i,J^{\pi_i}_{M^*})\}_{i=1}^{t}$ denote the sampling history up to the $t$-th iteration. To determine the $(t+1)$-th sample, we take any off-the-shelf AF, compute $\Psi(M'; \cH_t)$ for each model $M'\in \cM$, and then select the one with the largest AF value, \ie $M_{t+1}\in \arg\max_{M'\in \cM} \Psi(M'; \cH_t)$. 
    In our experiments, we use the celebrated GP-UCB \citep{srinivas2012information} as the AF.
    Then, upon sampling, we learn a policy $\pi_{t+1}$ based on the chosen dynamics model $M_{t+1}$ (\eg by applying SAC), and the function value $J^{\pi_{t+1}}_{M^*}$ can be obtained and included in the history. In practice, we use the Monte-Carlo estimates $R_{t+1}$ an an approximation of the true $J^{\pi_{t+1}}_{M^*}$.
    \item \textit{GP function prior}:  As in standard BO, here we impose a GP function prior that implicitly captures the structural properties of the objective function. Notably, GP serves as a surrogate model for facilitating probabilistic inference on the unknown function values. Specifically, under a GP prior, we assume that for any subset of models $\cM^\dagger \subseteq \cM$, their function values follow a multivariate normal distribution with mean and covariance characterized by a mean function $m:\cM\rightarrow \mathbb{R}$ and a covariance function $k:\cM\times \cM \rightarrow \mathbb{R}$. As in the standard BO literature, we simply take $m$ as a zero function. Recall that $R_{t}$ denotes the Monte-Carlo estimate of the true $J^{\pi_{t}}_{M^*}$. Let $\bm{R}_t$ denote the vector of all $R_i$, for $i\in \{1,\cdots,t\}$. For convenience, we let $\cM_t$ denote the set of models selected in the first $t$ iterations. 
    For any pair of model subset $\cM',\cM''$, define $\bm{K}(\cM',\cM'')$ to be a $\lvert \cM'\rvert \times \lvert \cM''\rvert$ covariance matrix, where the entries are the covariances $k(M',M'')$ of $M'\in \cM'$ and $M''\in \cM''$.
    Then, given the history $\cH_t$, the posterior distribution of $J^{\pi}_{M^*}$ of each model $M\in \cM$ follows a normal distribution $\mathcal{N}(\mu_t(M),\sigma^2_t(M))$, where
    \begin{align}
    \mu_t(M) &= {\bm{K}(M, \cM_t)} \bm{K}(\cM_t,\cM_t)^{-1} \bm{R}_t,   \\
    \begin{split}    
    \sigma^2_t(M) &= \bm{K}(M, M) - {\bm{K}(M,\cM_t)} \bm{K}(\cM_t,\cM_t)^{-1} \bm{K}(\cM_t, M).
    \end{split}
    \end{align}
    \item \textit{GP kernels for the covariance function}: To specify the covariance function $k(\cdot,\cdot)$ defined above, we adopt the common practice in BO and use a kernel function. For each pair of models $M',M''\in \cM$, under some pre-configured distance $d:\cM\times\cM\rightarrow [0,\infty)$, we use the popular Radial Basis Function (RBF) kernel \citep{williams2006gaussian}
    \begin{equation}
    k(M', M'') := \exp(-\frac{d(M', M'')^\text{2}}{\text{2}\ell^\text{2}}),\label{eq:RBF}
    \end{equation}
    where $\ell$ is the kernel lengthscale and ${d(M', M'')}$ is the model distance between $M'$ and $M''$.
    The overall procedure of BOMS is summarized in Algorithm \ref{algo:boms}. 
\end{itemize}

\begin{algorithm}[ht]
\caption{BOMS}\label{algo:boms}
\begin{algorithmic}
\Require True environment $M^*$, candidate model set $\cM$, and total number of selection iterations $T$.
\For{$t \gets 1$ to $T$} \Comment{\revECML{$t$ denotes the number of BO selection iterations}}
    \If{$t=1$} 
        \State Randomly choose a model $M_1$ from $\cM$. 
    \Else
        \State \multiline{Update the posterior $\mathcal{N}(\mu_t(M),\sigma^2_t(M))$ of each dynamics model $M\in \cM$.}
        \State 
            \multiline{Select ${M_{t}}\in \arg\max_{M\in \cM}\Psi(M;\cH_t)$.}
    \EndIf
    \State 
        \multiline{Learn a policy ${\pi_{t}}$ on the selected model ${M_{t}}$.}
    \State 
        \multiline{Evaluate $\pi_t$ by an estimate of empirical total reward $R_t$ based on online interactions with $M^*$.}
    \State 
        \multiline{Calculate the model distance $d(M_{t},M')$ between ${M_{t}}$ and all other models $M'\in \cM$.}
\EndFor
\State Return the model ${M}_{t^*}$ with $t^*:=\arg\max_{1\leq t\leq T}R_t$.

\end{algorithmic}
\end{algorithm}

\noindent \textbf{Remarks on realism of obtaining a candidate model set $\cM$:} We can naturally obtain $\cM$ by collecting the dynamics models learned during training under any offline MBRL method, without any additional training overhead. For example, in Section \ref{sec:exp}, we follow the model training procedure of MOPO and take the last 50 models as candidate models. This also induces a fair comparison between MOPO and BOMS as they both see the same group of dynamics models. 

\vspace{2mm}

\noindent \revECML{\textbf{Remarks on BOMS in safety-critical settings:} BOMS does not by itself guarantee safe execution of the queried policies. In safety-critical settings, BOMS can be directly combined with existing safety layers in RL, such as constrained policy execution, action shielding, early termination criteria, or a conservative fallback or behavior policy~\citep{achiam2017constrained,yu2022towards,thumm2022provably}. In this way, developing a constrained or risk-sensitive version of BOMS is a natural extension.}



\subsection{Model-Induced Kernels for BOMS}
\label{sec:boms:kernel}
To enable BOMS, one major challenge is to measure the distance $d(M',M'')$ between the dynamics models required by the kernel function. However, in offline MBRL, it remains unknown how to characterize the distance between dynamics models in a way that nicely reflects the smoothness in the total reward. 

\noindent{\textbf{Theoretical insights for BOMS kernel design:} To tackle the above challenge, we provide useful theoretical results that can motivate the subsequent design of a distance measure for offline MBRL. For convenience, as all the MDPs considered in this paper share the same $\cS, \cA, \omega$, and $\gamma$, in the sequel, we use a two-tuple $(P,r)$ to denote a model $M$, with a slight abuse of notation. Recall that $\pi_\beta$ denotes the behavior policy of the offline dataset. Here we use MOPO as an example and consider the penalized MDPs  $\widetilde{M}=({P},\widetilde{r})$, where $\widetilde{r}$ is the reward function penalized by the uncertainty estimator $u(s,a)$ and denoted as $\widetilde{r}(s,a) = r(s,a) - \lambda u(s,a)$ by MOPO, as described in Section \ref{sec:prelim}. 
To establish the following proposition, we make one regularity assumption that the value functions of interest are Lipschitz continuous in state, \ie there exist $L>0$ such that $\lvert V^{\pi}_{M}(s)-V^{\pi}_{M}(s)\rvert \leq L\cdot \lVert s-s'\lVert$, for all $s,s'\in \cS$, and ${L}$ is the Lipschitz constant.
We state Proposition~\ref{prop:model_dis} below, and the proof  is in Appendix \ref{app:proof}.}
\begin{restatable}{prop}{difprop}
\label{prop:model_dis}
Given two policies $\widetilde{\pi}_1$ and $\widetilde{\pi}_2$ which are the optimal policies learned on models $\widetilde{M}_1=(P_1,r_1)$ and $\widetilde{M}_2=(P_2,r_2)$, respectively. Then, we have 
\begin{align*}
    J^{\widetilde{\pi}_1}_{M^*}(\omega) - J^{\widetilde{\pi}_2}_{M^*}(\omega) \le &\Big(J^{\widetilde{\pi}_1}_{M^*}(\omega)-J^{\pi_\beta}_{M^*}(\omega) \Big)+2\lambda \epsilon(\pi_\beta) \\
    &+\frac{1}{1-\gamma} {\mathbb{E}} \Big[ \gamma L\norm\big{s'_1 - s'_2}_1 + \big| r_1(s,a)-r_2(s,a)\big| \Big],
\end{align*}
where the expectation is taken over $s\sim\omega$, $a\sim\widetilde{\pi}_1$, $s'_1\sim {P_1}(\cdot\rvert s,a)$, and $s'_2\sim {P_2}(\cdot \rvert s,a)$ and $\epsilon(\pi_\beta):= \mathbb{E}_{(s,a)\sim \rho^{\pi_\beta}}[u(s,a)]$ is the modeling error under behavior policy $\pi_\beta$.
\end{restatable}

\noindent\textbf{Insights offered by Proposition \ref{prop:model_dis}:} Notably, Proposition~\ref{prop:model_dis} shows that the policy performance gap of the learned models evaluated in the true environment is bounded by three main factors: 
(i) The first term is the expected discounted return difference between a learned policy and the behavior policy $\pi_\beta$. This term can be considered fixed since it does not change when we measure the distances between the selected model ${M}_t$ and other candidate models.
(ii) The second term is the modeling error along trajectories generated by $\pi_\beta$. Under MOPO, $\epsilon_u(\pi_\beta)$ should be quite small as $u(s,a)$ mainly estimates the discrepancy between the learned transition $\hat{P}$ and the true transition ${P}$, where $\hat{P}$ is learned from the dataset yielded from $\pi_\beta$. Therefore, for state-action pairs from $\rho^{\pi_\beta}$, $\hat{P}$ is close to the true transition ${P}$, and thus $\epsilon_u(\pi_\beta)$ should be relatively small. 
(iii) The third term is the next-step predictions difference between two dynamics models given $\widetilde{\pi}_1$. 
Moreover, note that the first two terms depend on the behavior policy, which is completely out of control by us, and the third term is the only term that reflects the discrepancy between $\widetilde{M_1}$ and $\widetilde{M}_2$. As a result, only the third term matters in measuring of the performance gap between dynamics models in BOMS.

\noindent\textbf{Model-induced kernels:} Built on the BOMS framework in Section \ref{sec:boms:framework} and the theoretical grounding provided above, we are ready to present the kernel used in BOMS. Recall from (\ref{eq:RBF}) that we use an RBF kernel in the design of the covariance matrix. To leverage Proposition \ref{prop:model_dis} in the context of BOMS, we take ${\pi}_1$ as the policy learned from the selected model $M_t=(P_t,r_t)$ at the $t$-th iteration, and ${\pi}_2$ be the policy from another candidate model $M=(P,r)$ in $\cM$. Then, motivated by Proposition \ref{prop:model_dis}, we design the model distance
\begin{equation}
    d(M_t, M):= \displaystyle\mathop{\mathbb{E}}\bigg[\lVert s_1'-s_2'\rVert+\alpha \big\lvert r_t(s,a)-r(s,a)\big\rvert\bigg],\label{eq:distance}
\end{equation} 
where the expectation is taken over $s\sim \cD_{\text{off}}$, $a \sim \pi_t(\cdot\rvert s)$, $s_1'\sim P_t(\cdot\rvert s,a)$, and $s_2'\sim P(\cdot\rvert s,a)$, and $\alpha$ is a parameter that balances the discrepancies in states and rewards. In the experiments, we find that $\alpha=1$ is generally a good choice.
In practice, we randomly sample a batch of states from the fixed dataset $\cD_{\text{off}}$ to obtain empirical estimates of the above distance $d(M_t, M)$. Additionally, since the covariance function $k(\cdot,\cdot)$ in BO is typically symmetric, we further impose a condition that the model distance $d$ enjoys symmetry, \ie ${d(M_t, M_j)}$ is equal to ${d(M_j, M_t)}$.

\vspace{2mm}
\noindent\textbf{Complexity of the proposed kernel:} Under BOMS with the proposed kernel and $N$ candidate models, the calculation of the covariance terms at the $t$-th iteration takes $O(t^2+tN)$, which is the same as standard BO. Hence, the proposed kernel does not introduce additional computational overhead.

\vspace{2mm}

\noindent \revECML{\textbf{Scalability of BOMS with candidate set size $N$:} Recall that $N$ is the number of candidate dynamics models and $T$ is the total number of BO iterations. Since BOMS targets small online evaluation budgets, $T$ is typically much smaller than $N$ (e.g., $T$ is fixed to 20 in our experiments). GP inference costs $O(T^3)$ for updating the posterior over evaluated models and $O(NT^2)$ for computing posterior statistics over all candidates, so it scales linearly with $N$ and can be parallelized across candidates. For the model-induced kernel, although precomputing all pairwise distances costs $O(N^2)$, BOMS only needs distances between the evaluated models and all candidates. Thus, distances can be computed on demand with $O(NT)$ evaluations over the full search, also in parallel. For very large candidate sets, BOMS can be combined with pre-screening, clustering, sparse GP approximations, or multi-stage search.}

\section{Experiments}
\label{sec:exp}

\subsection{Experimental Setup}

\noindent\textbf{Evaluation domains.} We evaluate BOMS on various benchmark RL tasks as follows:
\begin{itemize}[leftmargin=*]
    \item \textit{Locomotion tasks in MuJoCo}: We consider walker2d, hopper, and halfcheetah with 3 different types of datasets from D4RL \citep{fu2020d4rl} (namely, medium, medium-replay, medium-expert) generated from different behavior policies. These tasks are popular choices for offline RL evaluation due to the complex legged locomotion and challenges related to stability and control. Each dataset contains $10^6$ sampled transitions by default.
    \item \textit{Robot arm manipulation in Adroit and Meta-World}: We take the Adroit-pen task from D4RL, which involves the control of a 24-DoF simulated Shadow Hand robot tasked with placing the pen in a certain direction, with three different datasets (cloned, expert, mixed) generated from different behavior policies. The first two datasets are directly from D4RL, and the pen-mixed is a custom hybrid dataset with 50-50 split between the cloned and expert datasets. Each dataset contains $5\times 10^5$ transitions from the environment. Moreover, we take the door-open task in Meta-World \citep{yu2020meta}, which is goal-oriented and involves opening a door with a revolving joint under randomized door positions. Since the default Meta-World does not provide an offline dataset, we follow a data collection process similar to D4RL to obtain the offline medium-expert dataset.
\end{itemize}

\noindent\textbf{Candidate dynamics models.} Regarding training the candidate dynamics models, we follow the implementation steps outlined in MOPO \citep{yu2020mopo}. 
Specifically, we employ model ensembles by aggregating multiple independently trained neural network models to generate the transition function in the form of Gaussian distributions. 
{To acquire the candidate model set $\cM$, we construct a collection of 150 dynamics models for each task. Specifically, we obtain 50 models under each of the 3 distinct random seeds for each MuJoCo and Meta-World task, and we obtain 10 models under each of the 15 distinct seeds for the Adroit task.
For the calculation of model distance required by the GP covariance, BOMS randomly samples 1000 states from the offline dataset, obtains actions from the learned policy, and compares the predictions generated from different models.}

\noindent\textbf{Model selection process and inference regret.} In the model selection phase, we set the total number of BO iterations $T=20$. At each BO iteration, the evaluation of a policy is based on the observed Monte-Carlo returns averaged over 5 trajectories for MuJoCo and 20 trajectories for Adroit (note that this difference arises since the maximum trajectory lengths in MuJoCo and Adroit are 1000 and 100, respectively). For each selected dynamics model, we train the corresponding policy by employing SAC \citep{haarnoja2018soft} on the uncertainty-penalized MDP constructed by MOPO. For a more reliable performance evaluation, we conduct the entire model selection process for 10 trials and report the mean and standard deviation of the \textit{inference regret} defined as $\cR(t):=J^* - J^{\pi_{\text{out},t}}_{M^*}$, where ${J^*}$ denotes the true optimal total expected return among all model candidates in $\cM$ and $J^{\pi_{\text{out},t}}_{M^*}$ denotes the true total expected return of the policy output by the model selection algorithm at the end of $t$-th iteration. 

\revECML{All model-selection procedures are repeated over 10 trials to capture variability due to warm-start selection, Monte-Carlo evaluation noise, and policy training randomness.}

\vspace{2mm}

\noindent\textbf{Remark on the non-monotonicity of inference regret:} The inference regret defined above is a standard performance metric in the BO literature \citep{wang2017max,hvarfner2022joint,li2020multi}. Note that the inference regret $\mathcal{R}(t)$ is \textit{not} necessarily monotonically decreasing with $t$ since the policy $\pi_{\text{out},t}$ output by the model selection algorithm is not necessarily the best among those observed so far (due to the randomness in Monte-Carlo estimates).

\subsection{Results and Discussions}

\noindent\textbf{{Does BOMS outperform the existing model selection schemes?}}
To answer this, we compare BOMS with three baseline methods, including \textit{Validation}, \textit{OPE}, and \textit{Random Selection}. Validation is the default model selection scheme in MOPO, which chooses the model with the lowest validation loss in the model training process. Another baseline is OPE, which selects the dynamics model with the highest OPE value. As suggested by \citep{tang2021model}, we employ Fitted Q-Evaluation \citep{le2019batch} as our OPE method, which is a value-based algorithm that directly learns the Q-function of the policy. Random Selection selects uniformly at random the models to apply in the environment and outputs the model with the highest empirical average return at the end. 
\begin{table}[!ht]
\centering 

\caption{Normalized regrets of BOMS and the baselines. The best performance of each row is highlighted in bold. We use "med", "med-r", "mid-e" as the shorthands for medium, medium-replay, and medium-expert.}
\label{tab:mopo_regrets}
\begin{tabular}{c|l|c|c|c|c|c}
    \toprule
    \multicolumn{2}{c|}{\multirow{2}{*}{\vspace{-0.5em}Tasks}} & \multicolumn{3}{c|}{BOMS} &
    \multirow{2}{*}{\vspace{-0.5em}MOPO} &
    \multirow{2}{*}{\vspace{-0.5em}OPE} \\
    \cmidrule{3-5}
    \multicolumn{2}{c|}{} & {$T=5$} & {$T=10$} & {$T=20$} & & \\
    \midrule
    & med & 2.6$\pm$0.7 & 2.2$\pm$0.2 & \textbf{1.9$\pm$0.0} & 99.9 & 7.5\\
    walker2d & med-r & 69.9$\pm$17.7 & 44.0$\pm$26.7 & \textbf{16.9$\pm$17.5} & 77.1 & 94.3\\
    & med-e & 91.8$\pm$4.8 & 86.2$\pm$13.0 & 74.3$\pm$24.9 & 99.9 & \textbf{19.4}\\
    \midrule
    & med & 27.0$\pm$11.3 & 20.6$\pm$2.6\phantom{0} & \textbf{20.4$\pm$2.8\phantom{0}} & 66.0 & 78.4\\
    hopper & med-r & 5.5$\pm$1.4 & 5.1$\pm$2.6 & 5.2$\pm$2.6 & \textbf{3.1} & 59.3\\
    & med-e & 40.8$\pm$20.6 & 31.4$\pm$20.7 &\textbf{13.1$\pm$20.1} & 27.6 & 59.9 \\
    \midrule
    & med & 2.9$\pm$0.4 & 2.7$\pm$0.4 & \textbf{2.4$\pm$0.4} & 6.0 & 36.8 \\
    halfcheetah & med-r & 6.2$\pm$1.5 & 5.9$\pm$1.2 & \textbf{5.8$\pm$1.3} & 9.6 & 6.9 \\
    & med-e & 3.0$\pm$3.6 & 1.6$\pm$0.5 & \textbf{1.4$\pm$0.7} & 29.6 & 17.2 \\
    \midrule
    \multicolumn{2}{c|}{\small\textbf{MuJoCo}} & 27.7$\pm$6.9\phantom{0} & 22.2$\pm$7.6\phantom{0} & \textbf{15.7$\pm$7.8\phantom{0}} & 46.5 & 42.2 \\
    \midrule
    & cloned & 66.5$\pm$15.1 & 62.0$\pm$23.8 & \textbf{52.5$\pm$25.8} & 62.0 & 74.0\\
    pen & mixed & 21.8$\pm$16.2 & 15.3$\pm$11.2 &\textbf{10.9$\pm$10.3} & 40.9 & 61.7 \\
    & expert & 36.3$\pm$12.1 & 30.4$\pm$20.3 &\textbf{16.1$\pm$14.2} & 18.6 & 64.9\\
    \midrule
    \multicolumn{2}{c|}{\small\textbf{Pen}} & 54.3$\pm$12.4 & 51.1$\pm$14.5 & \textbf{39.3$\pm$22.3} & 51.8 & 74.2 \\
    \bottomrule
\end{tabular}
\end{table}

\begin{table}[!ht]
\caption{Success rates and normalized regrets of BOMS and the baselines on the Meta-World door-open task.}
\label{tab:door_open_regrets}
\centering
\begin{tabular}{c|ccc|c|c}
    \toprule
    \multirow{2}{*}{\vspace{-0.5em}\textbf{Door Open}} &
    \multicolumn{3}{c|}{\textbf{BOMS}} &
    \multirow{2}{*}{\vspace{-0.5em}\textbf{MOPO}} &
    \multirow{2}{*}{\vspace{-0.5em}\textbf{OPE}} \\
    \cmidrule(lr){2-4}
     & $T=5$ & $T=10$ & $T=20$ & & \\
    \midrule
    Success Rate & 1.3$\pm$4.0 & 6.5$\pm$15.7 & \textbf{16.3$\pm$23.5} & 0.0 & 0.0\\
    Regrets & 57.7$\pm$2.6 & 51.5$\pm$17.3 & \textbf{39.6$\pm$25.9} & 66.7 & 86.4\\
    \bottomrule
\end{tabular}
\end{table}

\begin{figure*}[!htp]
    \centering
    \includegraphics[width=0.8\textwidth]{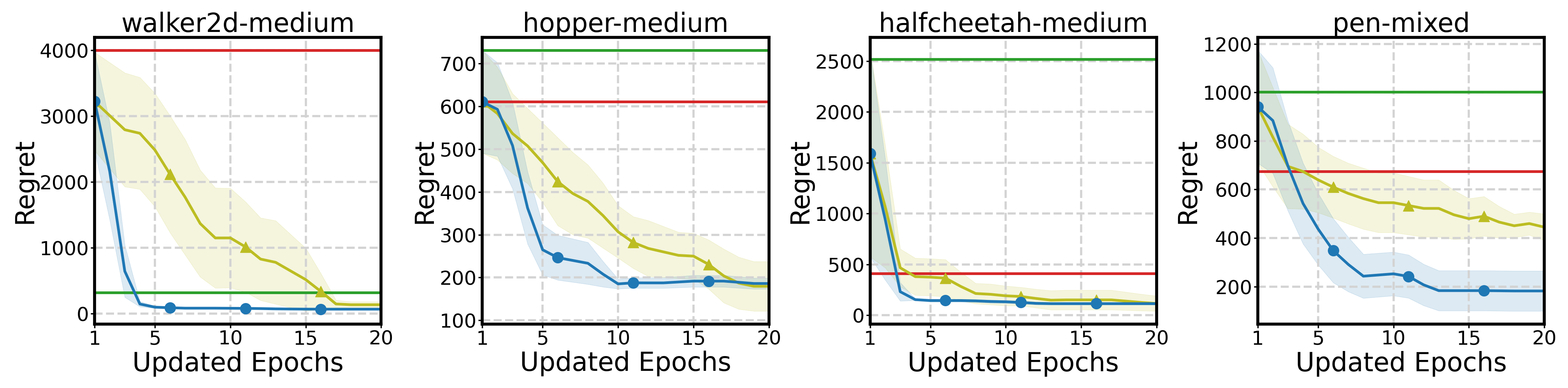}
    \includegraphics[width=0.6\textwidth]{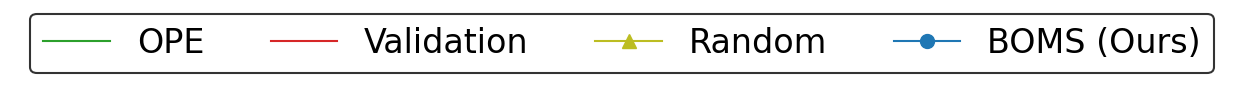}
    \caption{Comparison of BOMS and the baselines in inference regret. BOMS achieves 
    lower regrets than Validation and OPE in all the tasks after 5 iterations, which 
    correspond to only $1\%$-$2.5\%$ of the offline training data. \revECML{The x-axis denotes the number of BO selection iterations $t$ in Algorithm~\ref{algo:boms}.}}
    \label{fig:baselines}
\end{figure*}

Tables~\ref{tab:mopo_regrets}-\ref{tab:door_open_regrets} present the normalized regret scores of vanilla MOPO, OPE, and BOMS at the $5{\text{th}}$, $10{\text{th}}$, $20{\text{th}}$ iterations. The normalized regrets fall within the range of 0 to 100, where 0 indicates the regret of the optimal dynamics model in the task, and 100 indicates the return is 0. The normalized regrets indicate that under BOMS, it only takes about 5 selection iterations, which corresponds to only $1\%$-$2.5\%$ of training data, to significantly improve over vanilla MOPO and OPE in almost all tasks. \pch{Figure~\ref{fig:baselines} shows the regret curves. BOMS significantly outperforms Random Selection, and this further corroborates the benefit of using GP and the proposed model-induced kernel in capturing the similarity among the candidate dynamics models. Due to the page limit, more regret curves are in Figure~\ref{fig:main_appendix} in Appendix~\ref{app:exp}.}

\revECML{Note that the strong early performance of BOMS on walker2d-medium can be explained as follows: Since the first BOMS evaluation is random, the first-iteration result does not by itself indicate that BO has already learned an accurate surrogate. The improvement from the second iteration onward is more informative: in walker2d-medium, high-performing candidate models are concentrated in a relatively smooth neighborhood under the proposed model-induced distance, so GP-UCB can quickly exploit nearby models after observing the first return. This makes the early BO-guided selections more effective than uninformed random sampling.}

\begin{figure*}[!htp]
    \centering
    \includegraphics[width=0.8\textwidth]{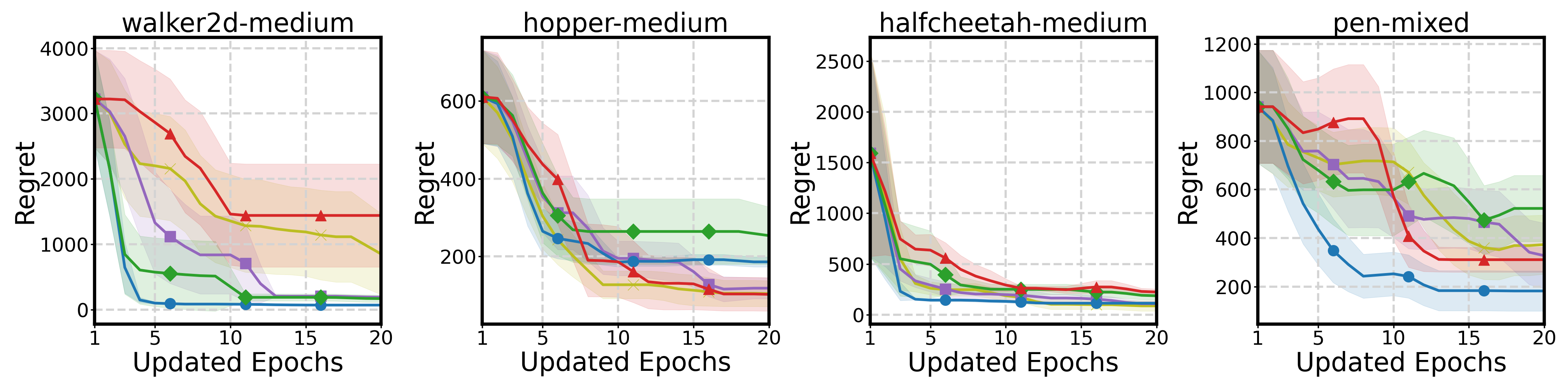}
    \includegraphics[width=0.6\textwidth]{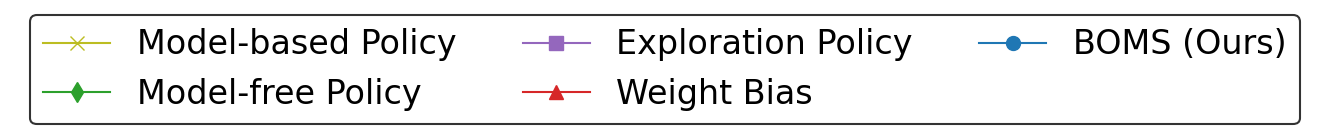}
    \caption{Comparison of various designs of model distance for BOMS in inference regret. These results corroborate the proposed model-induced kernel.  \revECML{The x-axis denotes the number of BO selection iterations $t$ in Algorithm~\ref{algo:boms}.}}
    \label{fig:ablations_selected}
\end{figure*}

\vspace{2mm}
\noindent\textbf{{Does BOMS effectively capture the relationship between models via the model-induced kernel?}}
To validate the proposed model-induced kernel, we further construct four other heuristic approaches for a comparison, namely \textit{Weight Bias}, \textit{Model-based Policy}, \textit{Model-free Policy}, and \textit{Exploratory Policy}. Weight Bias calculates the discrepancy between the weights and biases of the model neural networks as the model distance. The other three approaches generate rollouts with different policies (instead of the policy $\pi_t$ used in BOMS). Specifically, the Model-based Policy refers to the SAC policy trained with a pessimistic MDP, the Model-free Policy refers to the SAC policy learned from the pre-collected offline dataset, and the Exploratory Policy simply samples actions randomly. 
Figure~\ref{fig:ablations_selected} shows the evaluation results of BOMS under the proposed model distance and the above heuristics. We observe that BOMS indeed enjoys better regrets under the proposed model distance than other alternatives across almost all tasks. More regret curves are provided in Figure~\ref{fig:ablations} in Appendix \ref{app:exp}.

\begin{figure}[!ht]
    \centering
    \includegraphics[width=0.45\textwidth]{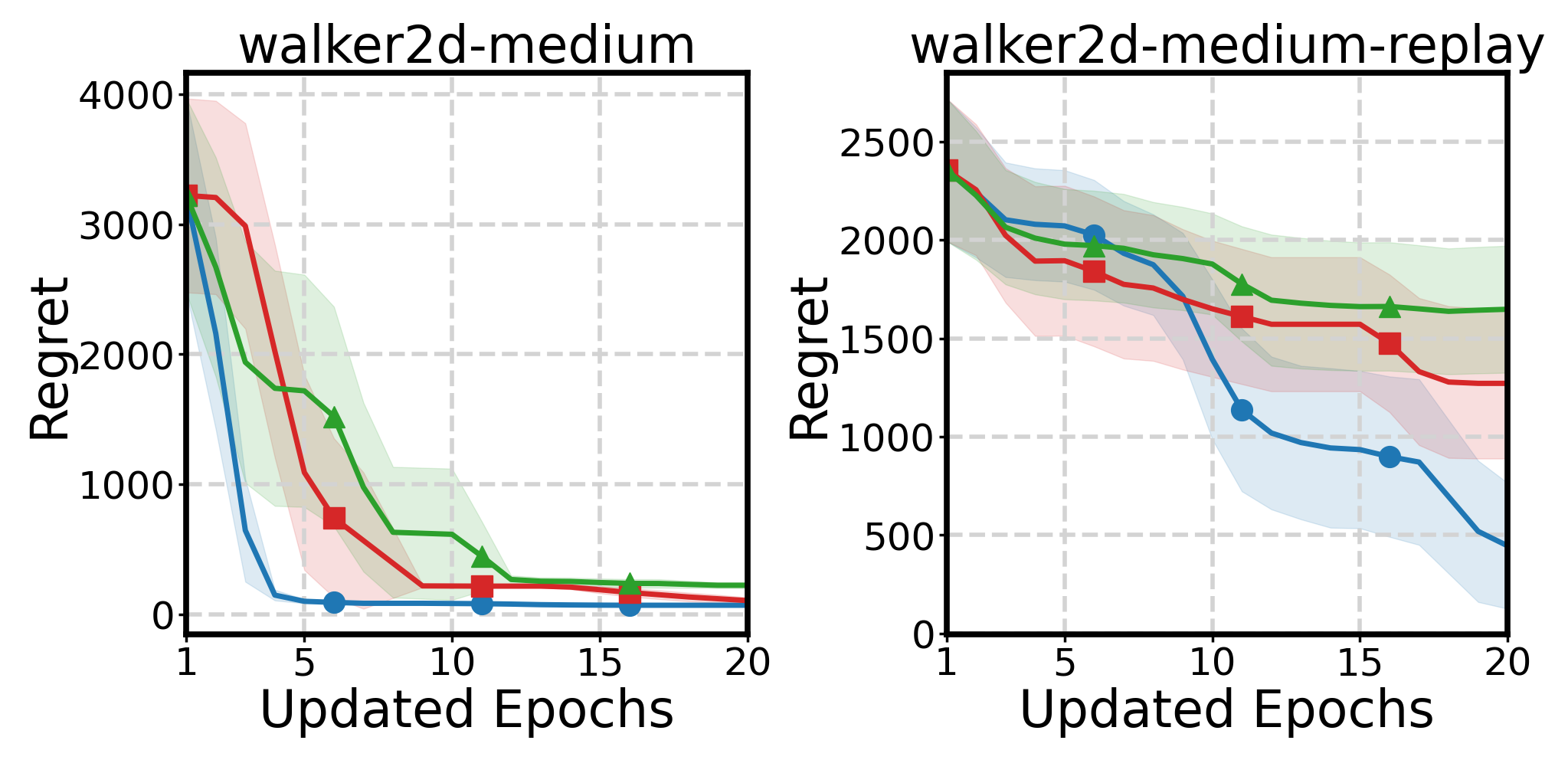} \\
    \includegraphics[width=0.45\textwidth]{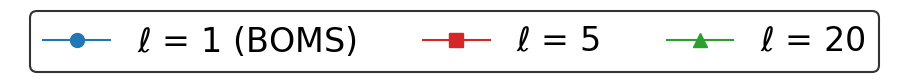}
    \caption{The inference regrets of BOMS with distance measure calculated under different rollout lengths.  \revECML{The x-axis denotes the number of BO selection iterations $t$ in Algorithm~\ref{algo:boms}.}}
    \label{fig:ablations_traj_len}
\end{figure}

On the other hand, recall from (\ref{eq:distance}) that by default BOMS takes only one-step predictions in measuring the model distance. One natural question to ask is whether \textit{multi-step predictions} generated by the learned models would help in the kernel design. To answer this, for the sake of comparison, we extend the model distance in (\ref{eq:distance}) to the multi-step version that takes the differences in states and rewards of the multi-step rollouts into account. Let $\ell$ denote the rollout length. From Figure~\ref{fig:ablations_traj_len}, we can see that BOMS under $\ell=1$ achieves the best regret performance, while the multi-step version under $\ell=5$, $20$ exhibit slower progress.
We hypothesize that this phenomenon is attributed to the inherent compounding modeling errors widely observed in MBRL.
\revECML{These results further highlight why we adopt the one-step model-induced distance in BOMS: It provides a more stable and policy-relevant signal while avoiding the compounding errors that can arise in longer model rollouts. Indeed, as also described in the celebrated work of MBPO~\citep{janner2019trust}, longer trajectory rollouts typically lead to more severe compounding errors in MBRL. At the same time, the effectiveness of this distance naturally depends on the quality of the learned dynamics models in the regions visited by candidate policies. When predictions become less reliable under substantial distribution shift, the induced similarity may become less precise. Developing more robust distance measures, for example by incorporating uncertainty estimates or representation-level comparisons, is an interesting direction for future work.}
\vspace{2mm}

\noindent\textbf{{Is BOMS generic such that it can be integrated with various offline MBRL methods?}}
We demonstrate the compatibility of BOMS with other offline MBRL methods by integrating BOMS with RAMBO \citep{rigter2022rambo}, another recent offline MBRL method.
Table~\ref{tab:rambo_regrets} shows the normalized regrets of BOMS-augmented RAMBO and the vanilla RAMBO. 
Similarly, BOMS-RAMBO can greatly enhance RAMBO within 10-20 iterations. This further validates the wide applicability of BOMS with offline MBRL in general. 

\begin{table}[!h]
\caption{Normalized regrets under vanilla RAMBO and BOMS-augmented RAMBO.}
\label{tab:rambo_regrets}
\centering
\begin{tabular}{c|c|ccc|c} 
\toprule
\multicolumn{2}{c|}{\multirow{2}{*}{Tasks}} & \multicolumn{3}{c|}{BOMS-Augmented RAMBO} & \multirow{2}{*}{RAMBO} \\
\cmidrule{3-5} 
\multicolumn{2}{c|}{} & $T=5$ & $T=10$ & $T=20$ & \\
\midrule
\multirow{3}{*}{walker2d} & med   & 79.3 $\pm$ 10.9 & 70.2 $\pm$ 24.7 & \textbf{37.7 $\pm$ 38.1} & 73.6 \\
                          & med-r & 93.0 $\pm$ 2.6  & 93.0 $\pm$ 2.5  & \textbf{92.7 $\pm$ 2.5}  & 99.3 \\
                          & med-e & 53.3 $\pm$ 10.6 & 43.0 $\pm$ 15.6 & \textbf{38.9 $\pm$ 17.2} & 65.4 \\
\bottomrule
\end{tabular}
\end{table}
\vspace{2mm}

\noindent\textbf{How does the choice of $\alpha$ in (\ref{eq:distance}) impact model selection?} Table~\ref{tab:rescale_reward} shows the results of BOMS with $\alpha \in \{0.1, 1, 10\}$. We observe that different values of $\alpha$ influence the selection of the models, though the influence is not significant. This also shows that $\alpha = 1$ is generally a good choice.

\begin{table}[!t]
\caption{The normalized regrets of BOMS with different $\alpha$ and the baselines.}
\label{tab:rescale_reward}
\begin{tabular}{c|l|r|c|c|c|c|c}
    \toprule
        \multicolumn{2}{c|}{\multirow{2}{*}{\vspace{-0.5em}Tasks}} & \multicolumn{1}{c|}{\multirow{2}{*}{\vspace{-0.5em}$\alpha$}} &
        \multicolumn{3}{c|}{BOMS} &
        \multirow{2}{*}{\vspace{-0.5em}MOPO} &
        \multirow{2}{*}{\vspace{-0.5em}OPE} \\
        \cmidrule{4-6}
        \multicolumn{2}{c|}{\multirow{2}{*}{}}&{\multirow{2}{*}{}}&{$T=5$} & {$T=10$} & {$T=20$} & & \\
    \midrule
         & med & 10.0 & 36.7$\pm$39.7 & 12.5$\pm$29.2 & \textbf{2.0$\pm$1.5} & 100.0 & 7.5 \\
         & med & 1.0 & 2.6$\pm$0.7 & 2.2$\pm$0.2 & \textbf{1.9$\pm$0.0} & 100.0 & 7.5\\
         & med & 0.1 & 61.6$\pm$47.0 & 17.9$\pm$25.3 & 8.6$\pm$7.8 & 100.0 & \textbf{7.5} \\
         & med-r & 10.0 & 62.3$\pm$28.6 & 53.5$\pm$23.3 & \textbf{38.4$\pm$21.8} & 77.1 & 94.3 \\
         walker2d & med-r & 1.0 & 69.9$\pm$17.7 & 44.0$\pm$26.7 & \textbf{17.0$\pm$17.5} & 77.1 & 94.3\\
         & med-r & 0.1 & 35.0 $\pm$27.2 & 17.2$\pm$0.1\phantom{0} & \textbf{17.2$\pm$0.1\phantom{0}} & 77.1 & 94.3 \\
         & med-e & 10.0 & 80.9$\pm$12.4 & 72.5$\pm$16.0 & 40.4$\pm$29.9 & 99.9 & \textbf{19.4} \\
         & med-e & 1.0 & 91.8$\pm$4.8\phantom{0} & 86.2$\pm$13.0 & 74.3$\pm$24.9 & 99.9 & \textbf{19.4}\\
         & med-e & 0.1 & 85.4$\pm$11.2 & 81.2$\pm$14.3 & 73.1$\pm$24.9 & 99.9 & \textbf{19.4} \\
    \midrule
         & med & 10.0 & 36.8$\pm$12.4 & 28.6$\pm$12.0 & \textbf{18.5$\pm$0.1\phantom{0}} & 66.0 & 78.4 \\
         & med & 1.0 & 27.0$\pm$11.3 & 20.6$\pm$2.6\phantom{0} & \textbf{20.4$\pm$2.8\phantom{0}} & 66.0 & 78.4\\
         & med & 0.1 & 20.0$\pm$12.3 & 17.5$\pm$11.8 & \textbf{16.7$\pm$11.4} & 66.0 & 78.4 \\
         & med-r & 10.0 & 5.1$\pm$3.1 & 5.4$\pm$2.9 & 5.5$\pm$1.1 & \textbf{3.1} & 59.3 \\
         hopper & med-r & 1.0 & 5.5$\pm$1.4 & 5.1$\pm$2.6 & 5.2$\pm$2.6 & \textbf{3.1} & 59.3\\
         & med-r & 0.1 & 15.1$\pm$21.8 & 5.9$\pm$2.8 & 3.9$\pm$2.9 & \textbf{3.1} & 59.3 \\
         & med-e & 10.0 & 36.3$\pm$18.1 & 28.8$\pm$13.2 & \textbf{22.1$\pm$15.5} & 27.6 & 59.9 \\
         & med-e & 1.0 & 40.8$\pm$20.6 & 31.4$\pm$20.7 & \textbf{13.1$\pm$20.1} & 27.6 & 59.9 \\
         & med-e & 0.1 & 44.2$\pm$8.5\phantom{0} & 40.3$\pm$4.2\phantom{0} & 40.3$\pm$4.2\phantom{0} & \textbf{27.6} & 59.9 \\
    \midrule
         & med & 10.0 & 4.5$\pm$2.5 & 3.0$\pm$1.6 & \textbf{2.6$\pm$1.4} & 6.0 & 36.8 \\
         & med & 1.0 & 2.9$\pm$0.4 & 2.7$\pm$0.4 & \textbf{2.4$\pm$0.4} & 6.0 & 36.8 \\
         & med & 0.1 & 5.4$\pm$5.4 & 2.6$\pm$1.6 & \textbf{1.9$\pm$0.4} & 6.0 & 36.8 \\
         & med-r & 10.0 & 8.5$\pm$2.1 & 7.7$\pm$1.7 & \textbf{6.0$\pm$1.5} & 9.6 & 6.9 \\
         halfcheetah & med-r & 1.0 & 6.2$\pm$1.5 & 5.9$\pm$1.2 & \textbf{5.8$\pm$1.3} & 9.6 & 6.9 \\
         & med-r & 0.1 & 8.2$\pm$1.9 & 7.6$\pm$1.8 & \textbf{6.7$\pm$1.5} & 9.6 & 6.9 \\
         & med-e & 10.0 & 16.0$\pm$18.4 & 8.5$\pm$5.0 & \textbf{5.8$\pm$4.0} & 29.6 & 17.2 \\
         & med-e & 1.0 & 3.0$\pm$3.6 & 1.6$\pm$0.5 & \textbf{1.4$\pm$0.7} & 29.6 & 17.2 \\
         & med-e & 0.1 & 8.2$\pm$4.6 & \textbf{7.1$\pm$4.4} & 8.3$\pm$3.6 & 29.6 & 17.2 \\

    \bottomrule
\end{tabular}
\end{table}
\section{Related Work}



One notable challenge of pure offline RL is that the absence of online interaction with the environment can severely limit the performance due to limited coverage of the fixed dataset. 
Indeed, the need for a small budget of online interactions has been studied and justified in the offline RL literature:

\noindent\textbf{Offline RL with policy selection.} Among these works, \citep{konyushova2021active} is the most relevant to ours in terms of problem formulation. Specifically, given a collection of candidate policies learned by any offline RL algorithm, \citep{konyushova2021active} proposes a setting called active offline policy selection (AOPS), which applies BO to actively choose a well-performing policy in the candidate policy set. Despite this high-level resemblance, AOPS does \textit{not} address the selection of dynamics models and is not readily applicable in active model selection. More specifically, the direct application of AOPS to our problem (\ie active model selection for offline MBRL) would \textit{require training policies for all the candidate dynamics models} at the first place, and this unavoidably leads to enormous computational overhead. Therefore, we view \citep{konyushova2021active} as an orthogonal direction to ours.

\noindent\textbf{Offline-to-online RL.} As a promising research direction, Offline-to-online RL (O2O) is meant to improve the policies learned by offline RL through fine-tuning on a small amount of online interaction.
To begin with, \citep{lee2022offline} introduces an innovative O2O approach that involves fine-tuning ensemble agents by combining offline and online transitions with a balanced replay scheme.
In a similar vein, \citep{agarwal2022reincarnating} presents Reincarnating RL (RRL), which aims to mitigate the data inefficiency of deep RL. Traditional RL algorithms typically start from scratch without leveraging any prior knowledge, resulting in computational and sample inefficiencies in practice. RRL reuses existing logged data or learned policies as an initialization for further real-world training to avoid redundant computations and improve the scalability.
More recently, the O2O problem has been studied from various perspectives, such as data mixing \citep{zheng2023adaptive,ball2023efficient}, policy fine-tuning \citep{xie2021policy,uchendu2023jump,zhang2023policy}, value-based fine-tuning \citep{zhang2024perspective}, and reward-based fine-tuning \citep{nair2023learning}. In contrast to the above works, BOMS utilizes online interactions to enhance the model selection for offline MBRL.

\pch{Moreover, due to the page limit, a review of the offline MBRL methods is provided in Appendix~\ref{app:related}.}

\section{Concluding Remarks}
In this paper, we propose BOMS, an active and sample-efficient model selection framework for offline MBRL, by recasting model selection as a BO problem. One critical novelty is the proposed model-induced kernel, which is theoretically grounded and enables GP posterior inference in BOMS.
Through extensive empirical evaluation, we validate several key design choices of BOMS and show that BOMS significantly outperforms the existing model selection methods through only a small amount of online interactions. 
We expect that BOMS can be integrated with various offline MBRL methods for reliable model selection.

\revECML{As BOMS is designed for settings where a small and controlled amount of online interaction is permitted, one inherent limitation is that BOMS is not directly applicable to strictly offline RL settings with zero online interaction. On the other hand, there are also several promising future directions: (1) As BOMS evaluates the performance during model selection, future work can study safer evaluation mechanisms, such as conservative policy screening, safety constraints, simulator-based pre-filtering, or risk-sensitive acquisition functions. (2) To further improve upon the one-step model-induced distance, future work could explore adaptive or representation-based distances that better capture policy-relevant model discrepancies. (3) Another interesting and practical direction is to scale BOMS to very large candidate pools, which can require sparse GP approximations and candidate pre-screening and clustering.}

\begin{appendices}


\section{Proof of Proposition \ref{prop:model_dis}}
\label{app:proof}


Before proving Proposition \ref{prop:model_dis}, we first introduce a useful property usually known as the \textit{simulation lemma} (\eg as indicated in Lemma A.1 in \citep{vemula2023virtues}).
\begin{lemma}[Simulation Lemma]
\label{lemma:sim}
    For any policy $\pi$, any initial state distribution $\omega$, and any pair of dynamics models $M', M''$, we have
    \begin{align}
        J^{\pi}_{M'}(\omega)-J^{\pi}_{M''}(\omega)
        &=\mathbb{E}_{s\sim \omega}[V^{\pi}_{M'}(s)-V^{\pi}_{M''}(s)]\\
        & = \frac{\gamma}{1-\gamma}\mathbb{E}_{s\sim\omega, a\sim\pi}\left [\mathbb{E}_{s'\sim M'(s,a)}[V^{\pi}_{M''}(s')]-\mathbb{E}_{s''\sim M''(s,a)}[V^{\pi}_{M''}(s'')]\right].
    \end{align}
\end{lemma}

For convenience, we restate the proposition as follows.

\begin{proof}
To begin with, we have
\begin{align}
    J^{\widetilde{\pi}_1}_{M^*}(\omega) - J^{\widetilde{\pi}_2}_{M^*}(\omega) 
    &= J^{\widetilde{\pi}_1}_{M^*}(\omega) - J^{\widetilde{\pi}_1}_{\widetilde{M_1}}(\omega) + J^{\widetilde{\pi}_2}_{\widetilde{M_2}}(\omega) - J^{\widetilde{\pi}_2}_{M^*}(\omega) +
     J^{\widetilde{\pi}_1}_{\widetilde{M_1}}(\omega) \underbrace{- J^{\widetilde{\pi}_2}_{\widetilde{M_2}}(\omega) + J^{\widetilde{\pi}_1}_{\widetilde{M}_2}(\omega)}_{\leq 0} - J^{\widetilde{\pi}_1}_{\widetilde{M}_2}(\omega) \\
    &\le \underbrace{\left(J^{\widetilde{\pi}_1}_{M^*}(\omega) - J^{\widetilde{\pi}_1}_{\widetilde{M}_1}(\omega)\right)}_{(A_1)}+ \underbrace{\left(J^{\widetilde{\pi}_2}_{\widetilde{M}_2}(\omega) - J^{\widetilde{\pi}_2}_{M^*}(\omega)\right)}_{(A_2)}
    + \underbrace{\left(J^{\widetilde{\pi}_1}_{\widetilde{M}_1}(\omega) - J^{\widetilde{\pi}_1}_{\widetilde{M}_2}(\omega)\right)}_{(A_3)},
\end{align}
where the first equality follows from the fact that $\widetilde{\pi}_1$ and $\widetilde{\pi}_2$ are optimal with respect to $\widetilde{M}_1$ and $\widetilde{M}_2$, respectively.
For the first and second terms on the RHS, we consult the theoretical results of MOPO. As $\widetilde{M}_1$ and $\widetilde{M}_2$ are uncertainty-penalized MDPs, one can verify that for any policy $\pi$,
\begin{align}
    J^{{\pi}}_{\widetilde{M}_1}(\omega) &\leq J^{{\pi}}_{M^*}(\omega),\\
    J^{{\pi}}_{\widetilde{M}_2}(\omega) &\leq J^{{\pi}}_{M^*}(\omega),
\end{align}
which have been shown in Equation (7) in MOPO \citep{yu2020mopo}.
Recall that $\widetilde{M}_1$ and $\widetilde{M}_2$ are the estimated dynamics models with the uncertainty penalty, and let $\widehat{M}_1$ be non-penalized version of $\widetilde{M}_1$. Then, we have 
\begin{align}
    (A_1)+(A_2) &=\left(J^{\widetilde{\pi}_1}_{M^*}(\omega) - J^{\widetilde{\pi}_1}_{\widetilde{M}_1}(\omega) \right)+ \left(J^{\widetilde{\pi}_2}_{\widetilde{M}_2}(\omega) - J^{\widetilde{\pi}_2}_{M^*}(\omega)\right) \\
    &= J^{\widetilde{\pi}_1}_{M^*}(\omega) - J^{{\pi}_\beta}_{M^*}(\omega) + \underbrace{J^{{\pi}_\beta}_{\widetilde{M}_1}(\omega) -J^{\widetilde{\pi}_1}_{\widetilde{M}_1}(\omega)}_{\leq 0} + J^{{\pi}_\beta}_{M^*}(\omega) - J^{{\pi}_\beta}_{\widetilde{M}_1}(\omega) + {J^{\widetilde{\pi}_2}_{\widetilde{M}_2}(\omega) - J^{\widetilde{\pi}_2}_{M^*}(\omega)} \\ 
    &\le J^{\widetilde{\pi}_1}_{M^*}(\omega) - J^{\pi_\beta}_{M^*}(\omega) + {J^{\pi_\beta}_{M^*}(\omega) - J^{{\pi}_\beta}_{\widetilde{M}_1}(\omega)} + \underbrace{J^{\widetilde{\pi}_2}_{\widetilde{M}_2}(\omega) - J^{\widetilde{\pi}_2}_{M^*}(\omega)}_{\leq 0} \\
    &\le J^{\widetilde{\pi}_1}_{M^*}(\omega) - J^{\pi_\beta}_{M^*}(\omega) + J^{\pi_\beta}_{M^*}(\omega) - J^{{\pi}_\beta}_{\widehat{M_1}}(\omega) + J^{{\pi}_\beta}_{\widehat{M_1}}(\omega) - J^{{\pi}_\beta}_{\widetilde{M}_1}(\omega) \\
    &\le J^{\widetilde{\pi}_1}_{M^*}(\omega) - J^{\pi_\beta}_{M^*}(\omega) + \left| J^{\pi_\beta}_{M^*}(\omega) - J^{{\pi}_\beta}_{\widehat{M_1}}(\omega) \right| +\lambda \epsilon_u (\pi_\beta) \\
    &\le \Big( J^{\widetilde{\pi}_1}_{M^*}(\omega)-J^{\pi_\beta}_{M^*}(\omega) \Big) + 2 \lambda \epsilon_u (\pi_\beta),
\end{align}
where the second last inequality holds by the definition of $\epsilon_u(\pi_\beta)$, and the last inequality holds by Lemma 4.1 in \citep{yu2020mopo}.

Next, to deal with the term $(A_3)$, we leverage the assumption that ${V^{\widetilde{\pi}_1}_{\widetilde{M}_1}(s)}$ is Lipschitz-continuous in state with a Lipschtize constant $L$. Take MuJoCo environments as an example, the state spaces of these tasks are continuous and composed of the positions of body parts and their corresponding velocities, and the reward functions are also calculated based on the body locations and velocities. In this scenario,  the value function is Lipschitz-continuous in states. 
Besides the assumption mentioned above, we also utilize the \textit{simulation lemma} in Lemma \ref{lemma:sim}. For notational convenience, below we let $\mathbb{E}$ be a shorthand for $\displaystyle {\mathbb{E}}_{\substack{\scriptstyle s\sim\omega, a\sim\widetilde{\pi}_1 \\ \scriptstyle s'_1\sim \widetilde{M_1}(\cdot\rvert s,a), s'_2\sim \widetilde{M_2}(\cdot \rvert s,a)}}$ throughout this proof. Then, an upper bound of $(A_3)$ is provided as follows:
\begin{align}
    J^{\widetilde{\pi}_1}_{\widetilde{M}_1}(\omega) - J^{\widetilde{\pi}_1}_{\widetilde{M}_2}(\omega)
    &= \displaystyle\mathop{\mathbb{E}}_{s\sim\omega} 
    {\left[ V^{\widetilde{\pi}_1}_{\widetilde{M}_1}(s) - V^{\widetilde{\pi}_1}_{\widetilde{M}_2}(s) \right]} \\
    &=\displaystyle\mathop{\mathbb{E}}_{s\sim\omega, a\sim\widetilde{\pi}_1} \Big[\Big(r_1(s,a)+ \gamma\displaystyle\mathop{\mathbb{E}}_{s'_1\sim \widetilde{M}_1(s,a)}[V^{\widetilde{\pi}_1}_{\widetilde{M}_1}(s'_1)]\Big) - \Big( r_2(s,a)+\gamma \displaystyle\mathop{\mathbb{E}}_{s'_2\sim \widetilde{M}_2(s,a)}[V^{\widetilde{\pi}_1}_{\widetilde{M}_2}(s'_2)] \Big)\Big] \\
    &= \mathop{\mathds{E}} \Big[\gamma \Big( V^{\widetilde{\pi}_1}_{\widetilde{M}_1}(s'_1)-V^{\widetilde{\pi}_1}_{\widetilde{M}_1}(s'_2)+V^{\widetilde{\pi}_1}_{\widetilde{M}_1}(s'_2)-V^{\widetilde{\pi}_1}_{\widetilde{M}_2}(s'_2) \Big) + \Big(r_1(s,a)-r_2(s,a)\Big) \Big]\\
    &= \frac{1}{1-\gamma} \displaystyle\mathop{\mathds{E}} {\left[ \gamma \Big(V^{\widetilde{\pi}_1}_{\widetilde{M}_1}(s'_1) - V^{\widetilde{\pi}_1}_{\widetilde{M}_1}(s'_2) \Big) + \Big(r_1(s,a)-r_2(s,a)\Big)\right]}\\
    & \tag*{\text{(by Lemma \ref{lemma:sim})}} \\
    &\le \frac{1}{1-\gamma}  \displaystyle\mathop{\mathds{E}} {\left[\gamma\left| V^{\widetilde{\pi}_1}_{\widetilde{M}_1}(s'_1) - V^{\widetilde{\pi}_1}_{\widetilde{M}_1}(s'_2)\right| + \Big| r_1(s,a)-r_2(s,a)\Big| \right]} \\
    &\le \frac{1}{1-\gamma}  \displaystyle\mathop{\mathds{E}} \Big[ \gamma L\norm\big{ s'_1 - s'_2}_1 + \big| r_1(s,a)-r_2(s,a)\big| \Big]
\end{align}

By combining all the upper bounds of ($A_1$)-($A_3$), we complete the proof of Proposition~\ref{prop:model_dis}.
\end{proof}

\newpage
\section{Experimental Configurations}
\label{app:config}

\noindent\textbf{Detailed configuration of Table~\ref{table:motivating}.} 
As a motivating experiment, we evaluate the validation-based and OPE-based model selection schemes on MOPO to highlight their limitations.
Regarding training the dynamics models and policies, we follow
the implementation steps outlined in MOPO: (i) Regarding the
dynamics models, we use model ensembles by aggregating multiple
independently trained neural network models to generate the
transition function in the form of Gaussian distributions. (ii) Regarding the policy policy, we use the popular SAC algorithm on the uncertainty-penalized MDP constructed by MOPO. To acquire the
candidate model set $\cM$, we create a collection of 150 dynamics models for each task (specifically, we obtain 50 models under each
of the 3 distinct random seeds for each MuJoCo locomotion task).
Regarding the OPE-based approach, we employ Fitted Q-Evaluation \citep{le2019batch} as our OPE method, which is a value-based algorithm that directly learns the Q-function of the policy.

\noindent\textbf{Hyperparameters of the experiments in Section~\ref{sec:exp}.}
Here we provide the hyperparameters utilized in our experiments. For the offline model-based RL setup, we directly adopt the default configuration of MOPO, including the hyperparameters in the dynamics model and policy training.
Regarding the calculation of model distance, we employ 1000 data points as input in BOMS. To maintain a comparable dataset size, we utilize 10 trajectories with a maximum of 100 rollout steps to assess model distance for Model-based Policy and Exploration Policy methods in the study.
For the online interaction with the environment, different environments lead to different budget settings. In MuJoCo tasks, we employ 5 trajectory samples for BO updates as the maximum rollout length is set to be 1000.
On the other hand, in the case of the Pen task in the Adroit environment, where the maximum number of steps is restricted to 100, we increase the number of trajectories to 20 (i.e., 2000 sample transitions) for the BO updates to get a comparable number of samples.

\noindent\textbf{Computing infrastructure:} All our experiments are conducted on machines with Intel Xeon Gold 6154 3.0GHz CPU, 90GB DDR4-2666 RDIMM memory, and NVIDIA Tesla V100 SXM2 GPUs.

\section{Additional Related Work}
\label{app:related}
The existing offline RL methods address the distribution shift problem from either a model-free or a model-based perspective:

\noindent\textbf{Offline model-free RL.} To tackle the distribution shift issue, several prior offline model-free RL approaches encode conservatism into the learned policies to prevent the learner from venturing into the out-of-distribution (OOD) areas. Specifically, conservatism can be implemented by either by either restricting the learned policy to stay aligned with the behavior policy \citep{fujimoto2019off, wu2019behavior, kumar2019stabilizing, nair2020awac, fujimoto2021minimalist, wang2022diffusion}, adding regularization to the value functions \citep{kumar2020conservative, kostrikov2021offline, bai2022pessimistic}, or applying penalties based on the uncertainty measures to construct conservative action value estimations \citep{an2021uncertainty, wu2021uncertainty, bai2022pessimistic}. Note that the above list is by no means exhaustive and is mainly meant to provide an overview of the main categories of offline model-free RL. Notably, among the above works, one idea in common is that these conservative model-free methods encourage the learned policy to align its actions with the patterns observed in the offline dataset.

\noindent\textbf{Offline model-based RL.} On the other hand, offline model-based RL \citep{janner2019trust, lu2021revisiting, argenson2020model, matsushima2020deployment} encode conservatism into the learned dynamics models based on the pre-collected dataset and thereafter optimize the policy. 
To begin with, MOPO \citep{yu2020mopo} and MOReL \citep{kidambi2020morel} are two seminal conservative offline model-based methods, which construct pessimistic MDPs by giving penalty to those OOD state-action pairs based on different uncertainty estimations.
Specifically, MOPO chooses the standard deviation of the transition distribution as the soft reward penalty, and MOReL introduces the ``halt state" in MDPs and imposes a fixed penalty when the total variation distance between the learned transition and the true transition exceeds a certain threshold.
Subsequently, \citep{matsushima2020deployment} proposes BREMEN, which  approaches the distributional shift problem through behavior-regularized model ensemble.
Moreover, given that uncertainty estimates with complex models can be unreliable, COMBO \citep{yu2021combo} addresses the distribution shift by directly regularizing the value function on out-of-support state-action tuples generated under the learned model, without requiring an explicit uncertainty estimation in the planning process. 
More recently, model-based RL with conservatism has been addressed from an adversarial learning perspective. For example, \citep{rigter2022rambo} proposes RAMBO-RL, which directly learns the models in an adversarial manner by enforcing conservatism through a minimax optimization problem over the policy and model spaces. Moreover, \citep{bhardwaj2023adversarial} proposes ARMOR, which adopts the concept of relative pessimism through a minimax problem over policies and models with respect to some reference policy. In this way, ARMOR can improve upon an arbitrary reference policy regardless of the quality of the offline dataset. 

{Despite the recent progress on offline model-based RL, it remains largely unexplored how to achieve effective model selection. The proposed BOMS serves as a practical solution that can complement the existing offline model-based RL literature.}

\section{Additional Experimental Results}
\label{app:exp}
\subsection{Comparison of Model Selection Methods}
\begin{figure*}[!htbp]
    \centering
    \includegraphics[width=0.9\textwidth]{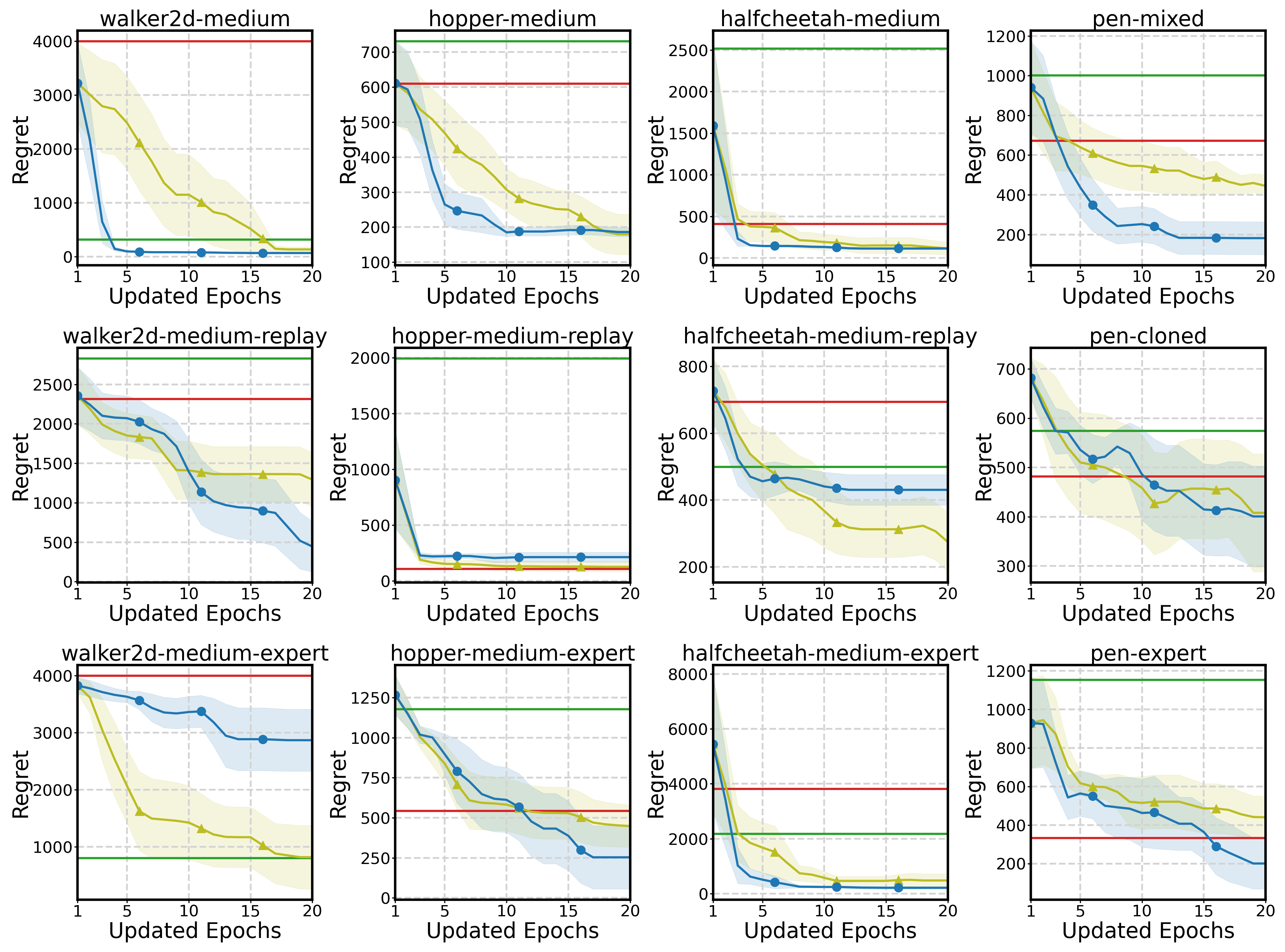}
    \includegraphics[width=0.9\textwidth]{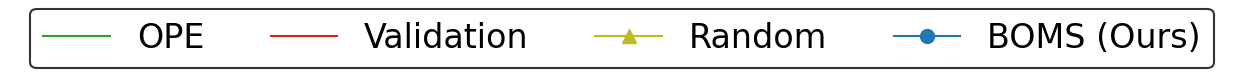}
    \caption{Comparison of BOMS and the baselines in inference regret. BOMS achieves lower regrets than Validation and OPE in almost all tasks after 5-10 iterations.}
    \label{fig:main_appendix}
\end{figure*}

Figure~\ref{fig:main_appendix} shows the regret performance of different model selection methods, highlighting the effectiveness of BOMS compared to baseline approaches.

\subsection{Comparison of Various Designs of Model Distance for BOMS in Inference Regret}
\begin{figure*}[!htbp]
    \centering
    \includegraphics[width=0.9\textwidth]{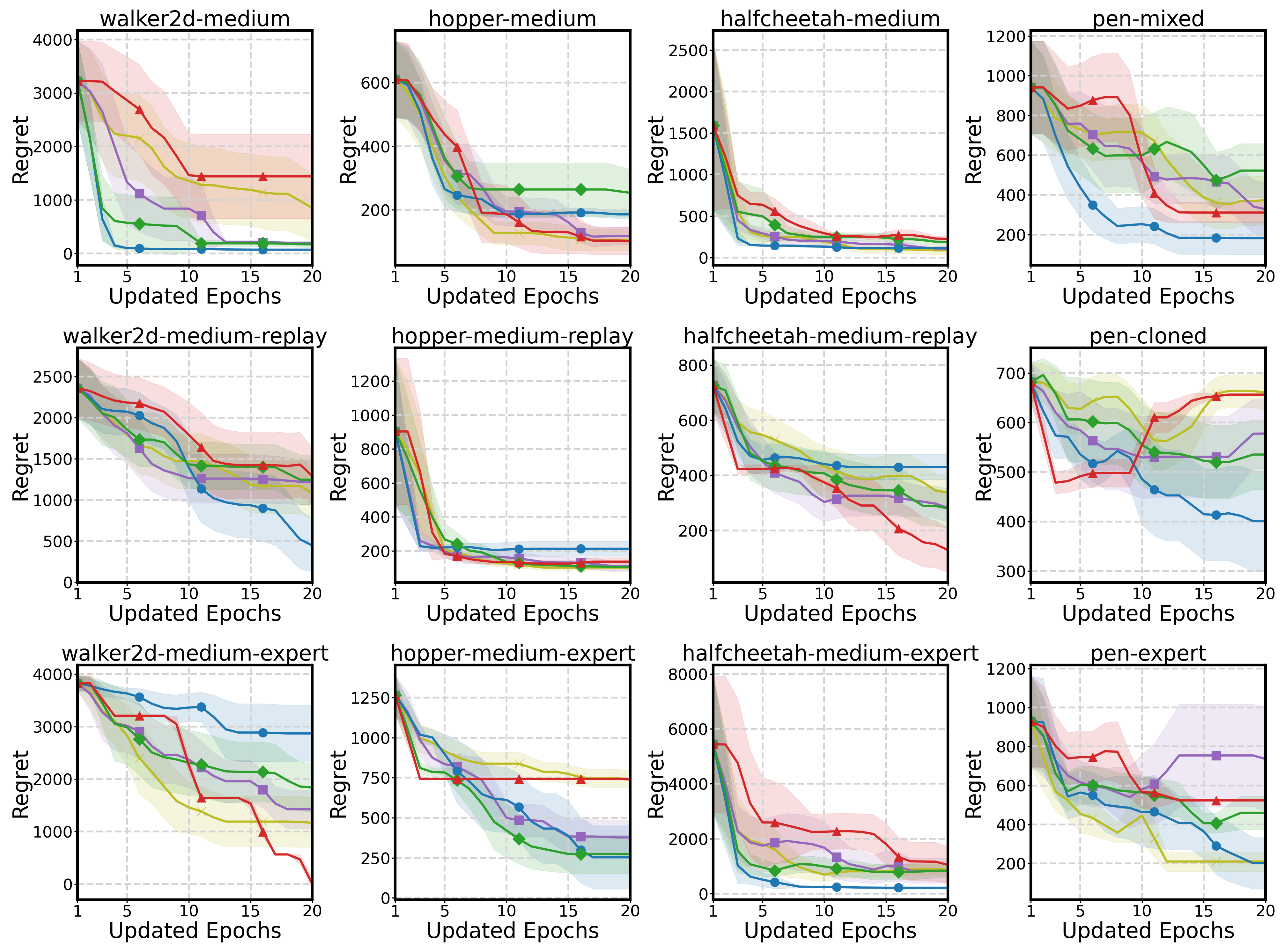}
    \includegraphics[width=0.9\textwidth]{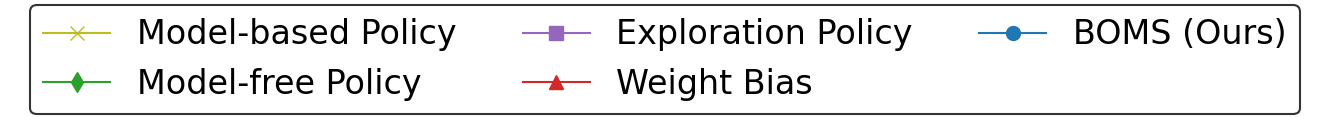}
    \caption{Comparison of various designs of model distance for BOMS in inference regret. These results corroborate the proposed model-induced kernel.}
    \label{fig:ablations}
\end{figure*}

{Figure~\ref{fig:ablations} shows the regret performance of BOMS under various designs of model distance. We can observe that BOMS with  the proposed distance defined in Equation (\ref{eq:distance}) generally achieves the lowest inference regret across all the tasks. This further corroborates the theoretically-grounded design suggested by Proposition \ref{prop:model_dis}.}

\newpage

\renewcommand{\arraystretch}{0.95}
\begin{table*}[!htbp]
\caption{Normalized rewards of our reproduced MOPO and the original MOPO.}
\label{tab:mopo_impl_comp}
\begin{adjustbox}{center}
\scalebox{0.85}{\begin{tabular}{c|l|c|c}
    \toprule
        \multicolumn{2}{c|}{Tasks} &
        {Our MOPO} &
        {Original MOPO} \\
    \midrule
        & med & -0.1 & \textbf{17.8}\\
        walker2d & med-r & 15.9 & \textbf{39.0}\\
        & med-e & 0.1 & \textbf{44.6}\\
    \midrule
        & med & 10.3 & \textbf{28.0}\\
        hopper & med-r & \textbf{99.2} & 67.5\\
        & med-e & \textbf{44.3} & 23.7\\
    \midrule
        & med & \textbf{54.6} & 42.3\\
        halfcheetah & med-r &  \textbf{54.7} & 53.1\\
        & med-e & \textbf{75.6} & 63.3\\
    \midrule
        & cloned & 6.9 & -\\
        pen & mixed & 28.7 & -\\
        & expert & 47.6 & -\\
    \bottomrule
\end{tabular}}
\end{adjustbox}
\end{table*}

\begin{sidewaystable*}[!ht]
\centering
\caption{Normalized rewards of BOMS and the offline RL benchmark methods. The best performance of each row is highlighted in bold.}
\label{tab:mopo_rewards}
\scalebox{0.85}{
\begin{tabular}{c|l|c|c|c|c|c|c|c|c|c|c}
\toprule
\multicolumn{2}{c|}{\multirow{2}{*}{\vspace{-0.5em}Tasks}}&
\multicolumn{4}{c|}{BOMS}&
\multirow{2}{*}{\vspace{-0.5em}MOPO}&
\multirow{2}{*}{\vspace{-0.5em}OPE}&
\multirow{2}{*}{\vspace{-0.5em}COMBO}&
\multirow{2}{*}{\vspace{-0.5em}IQL}&
\multirow{2}{*}{\vspace{-0.5em}TD3+BC}&
\multirow{2}{*}{\vspace{-0.5em}TT}\\
\cmidrule{3-6}
\multicolumn{2}{c|}{\multirow{2}{*}{}}&{$T=5$}&{$T=10$}&{$T=15$}&{$T=20$}& & & & & &\\
\midrule
&med&84.98$\pm$0.59\phantom{0}&85.34$\pm$0.13\phantom{0}&85.60$\pm$0.00&\textbf{85.60$\pm$0.00}&-0.02&80.72&63.76&62.64&66.96&65.04\\
walker2d&med-r&20.85$\pm$12.29&38.91$\pm$18.58&\phantom{0}45.59$\pm$17.18&\textbf{\phantom{0}57.68$\pm$12.19}&15.87&3.96&45.89&25.56&28.29&27.46\\
&med-e&7.81$\pm$4.52&13.13$\pm$12.39&\phantom{0}24.02$\pm$23.97&\phantom{0}24.40$\pm$23.64&0.07&76.66&\textbf{116.18}&111.82&112.33&92.83\\
\midrule
&med&21.32$\pm$3.21\phantom{0}&23.12$\pm$0.75\phantom{0}&23.00$\pm$0.84&23.17$\pm$0.79&10.26&6.74&88.82&89.68&80.27&\textbf{91.15}\\
hopper&med-r&96.74$\pm$1.37\phantom{0}&97.09$\pm$2.68\phantom{0}&97.05$\pm$2.69&97.05$\pm$2.69&\textbf{99.20}&42.02&70.07&38.21&24.80&40.08\\
&med-e&36.37$\pm$12.43&42.05$\pm$12.47&\phantom{0}50.31$\pm$13.13&\phantom{0}53.07$\pm$12.15&44.30&24.85&98.79&85.77&91.81&\textbf{99.26}\\
\midrule
&med&56.27$\pm$0.22\phantom{0}&56.37$\pm$0.24\phantom{0}&56.52$\pm$0.19&\textbf{56.52$\pm$0.19}&54.55&37.39&53.7&47.66&48.52&44.4\\
halfcheetah&med-r&56.65$\pm$0.86\phantom{0}&56.78$\pm$0.69\phantom{0}&56.86$\pm$0.74&\textbf{56.86$\pm$0.74}&54.66&56.24&54.66&44.94&45.33&44.85\\
&med-e&103.24$\pm$3.75\phantom{00}&104.63$\pm$0.56\phantom{00}&\phantom{0}104.87$\pm$0.74\phantom{0}&\textbf{104.87$\pm$0.74\phantom{0}}&75.57&88.44&89.9&59.18&61.81&29.05\\
\midrule
&cloned&5.69$\pm$4.01&6.91$\pm$6.33&\phantom{0}8.95$\pm$6.22&\textbf{\phantom{0}9.43$\pm$6.86}&6.89&3.70&-&-&-&-\\
pen&mixed&38.98$\pm$8.76\phantom{0}&42.49$\pm$6.04\phantom{0}&44.81$\pm$5.55&\textbf{44.85$\pm$5.55}&28.66&17.45&-&-&-&-\\
&expert&36.54$\pm$7.58\phantom{0}&40.19$\pm$12.65&45.06$\pm$9.59&\textbf{49.09$\pm$8.88}&47.58&18.67&-&-&-&-\\
\bottomrule
\end{tabular}
}
\end{sidewaystable*}

\subsection{Reproduction of MOPO}
Table~\ref{tab:mopo_impl_comp} shows the normalized rewards of our MOPO implementation and those of the original MOPO reported in \citep{yu2020mopo}.
Since our experiments are mainly based on MOPO, we would like to clarify that we have indeed tried our best on tuning MOPO properly, and hence our reproduced MOPO has similar or even better performance than the original MOPO results on halfcheetah and hopper. Additionally, although our MOPO perform on walker2d, after model selection with a small online interaction budget, our MOPO can still enjoy good results and even outperform other SOTA offline RL as shown in Table~\ref{tab:mopo_rewards}.

\subsection{Normalized Rewards Under BOMS and Other Offline RL Methods on D4RL }
Table~\ref{tab:mopo_rewards} shows the comparison of BOMS and other offline RL methods in terms of normalized rewards. Specifically, we include 4 popular benchmark methods, namely COMBO \citep{yu2021combo}, IQL \citep{kostrikov2021implicitq}, TD3+BC \citep{fujimoto2021minimalist}, and TT \citep{janner2021offline}, for comparison. We follow the default procedure provided by D4RL \citep{fu2020d4rl} to compute the normalized rewards: (i) A
normalized score of 0 corresponds to the average returns of a uniformly random policy; (ii) A normalized score of 100 corresponds to the average returns of a domain-specific expert policy. 
Notably, despite that the vanilla MOPO (with validation-based model selection) itself is not particularly strong, MOPO augmented by the model selection scheme of BOMS can outperform other benchmark offline RL methods on various offline RL tasks. This further showcases the practical values of the proposed BOMS in enhancing offline MBRL.

\end{appendices}
\clearpage
{
\setstretch{0.95}
\bibliography{ecml2026-BOMS}
}

\end{document}